\documentclass[10pt,journal,compsoc]{IEEEtran}
\usepackage[T1]{fontenc}
\usepackage[utf8]{inputenc}
\usepackage{graphicx}
\usepackage{booktabs}
\usepackage{multirow}
\usepackage{amsmath,amssymb,amsfonts}
\usepackage{xcolor}
\usepackage{algorithm}
\usepackage[noend]{algpseudocode}
\usepackage{url}
\usepackage[hidelinks]{hyperref}
\usepackage{enumitem}
\usepackage{array}
\usepackage{subcaption}
\setlist{nosep,leftmargin=*}

\newcommand{\system}{BatteryLake}

\begin{document}

\title{\system{}: Agentic, Physics-Grounded Curation of Heterogeneous Battery Aging Data and Benchmarking}

\author{Tianwen~Zhu,
        Hao~Wang,
        and~Yonggang~Wen,~\IEEEmembership{Fellow,~IEEE}%
\IEEEcompsocitemizethanks{\IEEEcompsocthanksitem T. Zhu, H. Wang, and Y. Wen are with the College of Computing and Data Science, Nanyang Technological University, Singapore.\protect\\
E-mail: \{tianwen001, hao-wang, ygwen\}@ntu.edu.sg}%
\thanks{}}%Manuscript received XX; revised XX.}}

\markboth{IEEE Transactions}%
{Zhu \MakeLowercase{\textit{et al.}}: \system{}: Agentic, Evidence-Grounded Curation of Heterogeneous Battery Aging Data}

\IEEEtitleabstractindextext{%
\begin{abstract}
Public battery aging datasets are a critical asset for advanced health management. However, their practical use is often limited by inconsistent file formats, unclear data schemas, and metadata that are dispersed across repositories an publications. Current curation practices remain largely manual and dataset-specific, making them difficult to reproduce. Meanwhile, general-purpose data integration tools often fail to capture the domain-specific semantics of electrochemical time-series data. In this paper, we presents \system{}, a governed data lakehouse that turns raw public battery data into benchmark-ready assets through an \emph{agentic, evidence-grounded} curation framework. We make three main contributions to the automated curation of public battery datasets.
The first capability combines evidence-grounded metadata extraction with explicit abstention, and schema mapping with program synthesis for dataset-specific converters. Both tasks are handled by large language model (LLM) agents. Each agent output must be grounded in verbatim source evidence. If such evidence is unavailable, the agent abstains rather than producing an unsupported result. 
Second, we design a human-in-the-loop verification mechanism that frames field extraction as a selective prediction problem, deferring low-confidence extractions to human review. Low-confidence fields are routed to domain experts, while high-confidence fields are accepted only when the residual error rate is bounded below a prescribed threshold. The admitted data must then pass a 26-rule validation gate covering schema consistency, statistical validity, and physical plausibility.
%Third, on top of the curated lake we release an open benchmark covering 12 fully curated datasets from 12 institutions ($\sim$720 cells, $\sim$323K charge--discharge cycles) with standardized state-of-health and remaining-useful-life tasks, three split protocols, and eight baseline model families.
Third, on top of the curated data lake we release an open benchmark spanning 41 curated datasets from over 25 institutions, covering various chemistries across cylindrical, pouch, and prismatic formats and spanning cycle-aging, drive-cycle, impedance, calendar-aging, and thermal-abuse regimes, with standardized state of health (SOH) and remaining useful life (RUL) tasks, three split protocols, and eight baseline model families.
% Experiments show that evidence grounding reduces hallucinated metadata from \xx\% to \xx\% while retaining \xx\% field coverage, agentic conversion succeeds on \xx\% of heterogeneous sources, and human curation effort drops by \xx$\times$ relative to fully manual pipelines. 
The platform, benchmark, and curation protocol are publicly available at \url{https://tianwen1209.github.io/batterylake/}.
\end{abstract}

}
 
\maketitle
\IEEEdisplaynontitleabstractindextext
\IEEEpeerreviewmaketitle
 
\section{Introduction}\label{sec:intro}
 
\IEEEPARstart{L}{ithium-ion} batteries are now widely deployed in electric vehicles, grid-scale storage, and portable electronics, and in all of these settings their capacity and safety margins decline as the cells age. How quickly a cell loses capacity, and how close it is to end of life, determines when it must be retired, how much range or backup it can still deliver, and whether it can be reused or must be replaced.
These quantities are hard to obtain from first principles, because degradation couples chemistry, temperature, load, and usage history, and no compact physical model captures all of these effects across cell types and operating conditions. As a result, degradation behavior is increasingly learned from data, using machine learning over large-scale cycling experiments~\cite{severson2019,attia2020}. Over the past two decades, laboratories worldwide have released dozens of aging datasets, from the NASA Prognostics Center of Excellence~\cite{saha2007} and CALCE~\cite{xing2013} to the 124-cell fast-charging study of Severson et al.~\cite{severson2019}, which together contain millions of charge--discharge cycles. In principle, this is ample data for state of health (SOH) estimation, remaining useful life (RUL) prediction, and cross-condition generalization research. In practice, however, its large-scale use is still limited by poor curation: surveys of open battery data report fragmented repositories, inconsistent file formats, and incomplete documentation~\cite{dosreis2021}.
 
The obstacle is not data \emph{volume} but data \emph{engineering}. Three structural problems recur. (i) \textbf{Format and schema heterogeneity.} Raw releases arrive in many file formats: CSV, Excel, MATLAB \texttt{.mat}, HDF5, JSON, Parquet, and proprietary cycler exports (Arbin, MACCOR, Neware). Within them, the same physical measurement appears under dozens of aliases (\texttt{Voltage}, \texttt{Ecell/V}, \texttt{V}), in inconsistent units (\texttt{Current\_mA} vs.\ amperes; Kelvin vs.\ Celsius), and with conflicting sign conventions for charge and discharge current. (ii) \textbf{Metadata dispersion.} Experiment-defining facts such as cell chemistry, nominal capacity, cycling protocol, temperature, cutoff voltages, and software license are rarely stored in the data files themselves. This information is usually scattered across repository landing pages, README files, and the prose of companion papers, and should be recovered by hand. (iii) \textbf{Benchmark non-reproducibility.} Different studies use different parsing routines, cycle segmentation methods, label definitions, and train/test splits.
As a result, published model comparisons are hard to reproduce and rarely directly comparable, as documented in the battery machine learning literature~\cite{zhang2024batteryml}.
 
Classical battery data integration offers only partial relief. Schema
matchers such as Cupid~\cite{madhavan2001} operate on relational
metadata and cannot exploit the physical semantics (value ranges,
units, monotonic degradation) that disambiguate battery signals; data
validation systems~\cite{schelter2018} check constraints but do not
author the mappings; and data lake management~\cite{nargesian2019}
presumes ingested data whose provenance is already understood. Recent
work shows that large language models can perform data wrangling and
extraction tasks with little supervision~\cite{narayan2022,
arora2023}, but large language models (LLMs) used naively for scientific curation introduce a
failure mode worse than missing data: \emph{plausible fabrication} of
metadata (e.g., guessing a chemistry from a dataset name), which
silently corrupts every downstream analysis.
 
We argue that trustworthy curation of scientific data at scale requires coupling LLM agents' flexibility
with three safeguards that are largely absent from prior LLM data
wrangling work: (1) \emph{evidence grounding with explicit
abstention}: every extracted data must be justified by verbatim
source text, and fields without support are marked ``not stated''
rather than imputed; (2) \emph{human verification}:
Per-field confidence scores route uncertain decisions to human experts and allow the review process to move beyond all-or-nothing manual inspection. By adjusting the confidence threshold, the system can trade automatic coverage against the residual error risk of admitted fields; and (3) \emph{executable
validation}: agent-synthesized converters are accepted only if
their outputs pass a machine-checkable gate combining schema,
statistical, and battery-specific physical-plausibility constraints.

We instantiate these principles in \system{}, a governed lakehouse
and public benchmark for battery aging data. This paper makes the following contributions:
\begin{itemize}
\item \textbf{Problem formalization.} We formalize battery
dataset onboarding as (a) evidence-grounded structured extraction
with abstention over multi-source documents, and (b) schema mapping
with converter program synthesis, and we cast human review as
selective prediction with an explicit risk--coverage objective.
\item \textbf{Agentic curation framework.} We design a two-stage,
provenance-ranked extraction algorithm over dataset landing pages and
companion papers, a file-role classifier and generate--validate--repair
loop for converter synthesis, and a 26-rule validation gate spanning
four quality dimensions including physical plausibility.
\item \textbf{Governed lakehouse and open benchmark.} We define a
four-layer canonical data model with typed units and integrity
constraints, and release a benchmark of 41 curated datasets from
25 institutions with standardized SOH/RUL tasks, three split
protocols, eight baseline families, and versioned reproducibility
manifests.
%\item \textbf{Evaluation framework.} We design a comprehensive evaluation spanning extraction fidelity, hallucination and abstention behavior, risk--coverage of the human gate, conversion success, and downstream benchmark utility, against rule-based and ungrounded-LLM baselines (Section~\ref{sec:eval}).
\end{itemize}
 
While \system{} is instantiated for batteries, the framework includes
evidence-grounded extraction, selective verification, and physically
validated program synthesis. We design \system{} to address a general pattern in scientific data engineering: converting long tails of small, heterogeneous, under-documented datasets into governed, analysis-ready collections.
 
\section{Related Work}\label{sec:related}
 
\textbf{LLMs for data integration and extraction.} Foundation models have recently been explored for data integration and preparation tasks. Prior work has shown their ability to perform entity matching, error detection, and imputation with few or no labeled examples~\cite{narayan2022}. Other studies have used them to synthesize extraction code for semi-structured corpora~\cite{arora2023} and to unify diverse data preparation tasks under instruction tuning~\cite{zhang2024jellyfish}. Pre-LLM neural
matchers such as Ditto~\cite{li2020ditto} and unified matching models
such as Unicorn~\cite{tu2023unicorn} established that pretrained language models transfer across integration tasks. First, we target \emph{scientific} metadata, whose ground truth is sparse and scattered across web pages and papers. For such data, we make abstention and verbatim evidence first-class outputs, rather than the calibration afterthought they are in prior data wrangling systems.
Our converter synthesis relates to agentic tool-use~\cite{yao2023react}, but adds a domain-specific executable validator as the acceptance test.
 
\textbf{Data lakes, validation, and governance.} Data lake management
research addresses ingestion, discovery, and versioning over schema-
on-read collections~\cite{nargesian2019}; lakehouse systems such as
Delta Lake add ACID guarantees over open formats~\cite{armbrust2020};
and declarative validation systems verify data quality constraints at
scale~\cite{schelter2018}. Constraint-driven repair engines such as
HoloClean~\cite{rekatsinas2017} clean data already inside the
warehouse. \system{} complements this line: it governs the
\emph{boundary} of the lake, using agents to author the mappings and
metadata that validation systems presuppose, and extends generic
constraints with electrochemical plausibility rules (voltage bounds,
coulombic-efficiency ranges, capacity monotonicity).
 
\textbf{Battery data infrastructure and benchmarks.} The battery
community has produced open datasets~\cite{saha2007, xing2013,
severson2019, birkl2017}, surveys of their fragmentation
\cite{dosreis2021}, processing toolkits such as BEEP for
cycler-file featurization~\cite{herring2020}, ontologies for
semantic interoperability~\cite{clark2022}, and benchmark libraries
such as BatteryML~\cite{zhang2024batteryml}. These efforts either
hand-curate a fixed set of datasets or standardize formats
prospectively for new data. \system{} is, to the best of our knowledge, the
first system to automate \emph{retrospective} onboarding of the long
tail of already-published battery datasets with auditable provenance,
and the first battery benchmark whose every admitted dataset carries
machine-checkable curation evidence, aligning the collection with
FAIR principles~\cite{wilkinson2016}.
 
% \textbf{Selective prediction and human-in-the-loop curation.} Our
% verification gate builds on selective classification, which trades
% coverage for accuracy by abstaining on low-confidence
% inputs~\cite{elyaniv2010, geifman2017}, and on the observation that
% modern neural predictors require calibration~\cite{guo2017}. We apply
% the risk--coverage lens not to model inference but to \emph{curation
% decisions}, deriving review budgets and residual-risk bounds for
% metadata admitted into a governed lake.
 
\section{Problem Formulation}\label{sec:problem}
 
We first define the canonical target of curation
(\S\ref{sec:model}), then the two onboarding problems
(\S\ref{sec:extraction-def}, \S\ref{sec:mapping-def}), and the
selective verification objective (\S\ref{sec:selective-def}).
 
\subsection{Canonical Data Model}\label{sec:model}
 
\system{} organizes each curated dataset as a four-layer relational
object $\mathcal{D} = (M_d, M_c, S, T)$:
\begin{itemize}
\item \emph{Dataset metadata} $M_d$: one tuple per dataset over
attribute set $F_d$ = \{chemistry, cathode\_material,
anode\_material, nominal\_capacity\_Ah, nominal\_voltage\_V,
temperature\_C, charge\_protocol, discharge\_protocol, C\_rate,
cutoff\_voltage\_upper, cutoff\_voltage\_lower, form\_factor,
brand\_or\_manufacturer, source\_url, paper\_url, license, \ldots\};
\item \emph{Cell metadata} $M_c$: one tuple per cell, keyed by
(\texttt{dataset\_id}, \texttt{cell\_id});
\item \emph{Cycle summary} $S$: one tuple per (cell, cycle) with
capacities, coulombic and energy efficiency, SOH, RUL, durations,
voltage/temperature statistics, and quality flags;
\item \emph{Time series} $T$: one tuple per sample with canonical
signals \texttt{time\_s}, \texttt{voltage\_V}, \texttt{current\_A},
\texttt{temperature\_C}, \texttt{charge/discharge\_capacity\_Ah},
\texttt{step\_type}, under fixed units and the sign convention
$I>0$ on charge.
\end{itemize}
Each attribute $a$ carries a type $\tau(a)$, a unit $u(a)$, and a
domain $\mathrm{dom}(a)$ (e.g., $\mathrm{dom}(\texttt{voltage\_V})
\subseteq [0, 6]$\,V for single cells). Layers are linked by foreign
keys $T \rightarrow S \rightarrow M_c \rightarrow M_d$, and a
constraint set $\Sigma$ of $|\Sigma| = 26$ rules governs admission
(Section~\ref{sec:gate}). Every dataset additionally receives a
machine-readable reference name
$\texttt{ref\_name} =
\texttt{YYYY\_SOURCE\_CHEM\_FORM\_CHRGC\_DCHRGC\_TEMP}$
that encodes its identity for cross-dataset filtering.
 
\subsection{Evidence-Grounded Extraction with Abstention}
\label{sec:extraction-def}
 
Let $F \subseteq F_d$ be the target metadata fields and let
$\mathcal{E} = \{E_1, \ldots, E_k\}$ be an ordered set of evidence
documents with provenance ranks (e.g., $E_1$ the dataset landing
page, $E_2$ the companion paper). An \emph{evidence-grounded
extractor} is a function
\begin{equation}
\Phi(F, \mathcal{E}) = \{(f,\; v_f,\; e_f,\; s_f,\; c_f)\}_{f \in F},
\end{equation}
where $v_f \in \mathrm{dom}(f) \cup \{\bot\}$ is the extracted value
or the abstention symbol $\bot$ (rendered as ``source page not
stated''), $e_f$ is a verbatim evidence span from some $E_i$, $s_f =
i$ records the provenance, and $c_f \in [0,1]$ is a confidence score.
The output must satisfy the \emph{grounding constraint}: if $v_f \neq
\bot$ then $e_f$ must occur in $E_{s_f}$ and entail $v_f$. Given a
gold annotation $v_f^\ast$ (which may itself be $\bot$ when no source
states the field), we distinguish four outcomes: correct extraction
($v_f = v_f^\ast \neq \bot$), incorrect extraction, correct
abstention ($v_f = v_f^\ast = \bot$), and \emph{hallucination}
($v_f \neq \bot$ while $v_f^\ast = \bot$), which is the failure mode our
design explicitly targets. The extractor should maximize coverage
$\mathbb{P}[v_f \neq \bot]$ subject to a bound on the conditional
error $\mathbb{P}[v_f \neq v_f^\ast \mid v_f \neq \bot]$.
 
\subsection{Schema Mapping and Converter Synthesis}
\label{sec:mapping-def}
 
A raw source is a file collection $R = \{r_1, \ldots, r_m\}$ with
unknown roles and layouts. Onboarding $R$ requires: (i) a
\emph{role labeling} $\rho: R \rightarrow
\{\textsf{measurement}, \textsf{metadata\_only}\}$ separating files
that can yield $T$ and $S$ from documentation usable only as
extraction evidence; (ii) an \emph{attribute mapping}
$h: A_R \rightarrow A_{\mathcal{D}} \cup \{\bot\}$ from raw columns,
MAT keys, or nested fields to canonical attributes, together with
unit conversions $g_a$ and sign normalization; and (iii) a
\emph{converter program} $\pi$ such that $\pi(R) = (\hat{T}, \hat{S},
\hat{M_c})$ and $\pi(R) \models \Sigma$. Because raw layouts are
open-ended, we do not fix a parser family; instead an LLM agent
\emph{synthesizes} $\pi$ per dataset, and correctness is enforced
extrinsically by an executable validator for $\Sigma$
(Section~\ref{sec:agentic}). The mapping $h$ must likewise be
grounded: the agent may only bind a raw label to a canonical signal
if header text, embedded units, or value statistics (range,
polarity, monotonicity) support the binding, and must abstain
otherwise.
 
\subsection{Selective Human Verification}\label{sec:selective-def}
 
Extraction and mapping decisions carry confidences $c$. A
\emph{verification policy} with threshold $\tau$ auto-accepts
decisions with $c \geq \tau$ and routes the rest to a human reviewer,
who edits or confirms them through the platform interface; no row
enters the lake unconfirmed. Let $\mathrm{cov}(\tau)$ be the fraction
auto-accepted and $\mathrm{risk}(\tau)$ the error rate among
auto-accepted decisions. Assuming reviewer decisions are correct,
the residual error of admitted metadata is
$\mathrm{risk}(\tau)\cdot\mathrm{cov}(\tau)$, and the expected human
effort is proportional to $1-\mathrm{cov}(\tau)$. The curation
operating point is chosen on the empirical risk--coverage curve
\cite{elyaniv2010,geifman2017} to meet a target residual risk
$\alpha$ with minimal review budget; we report the full curve rather
than a single point. This formulation turns
the ubiquitous but ad hoc ``human checks the AI'' practice into a
measurable, tunable component with an explicit quality guarantee.
 
\section{The \system{} Curation Framework}\label{sec:framework}
 
\begin{figure*}[t]
\centering
\includegraphics[width=0.98\textwidth]{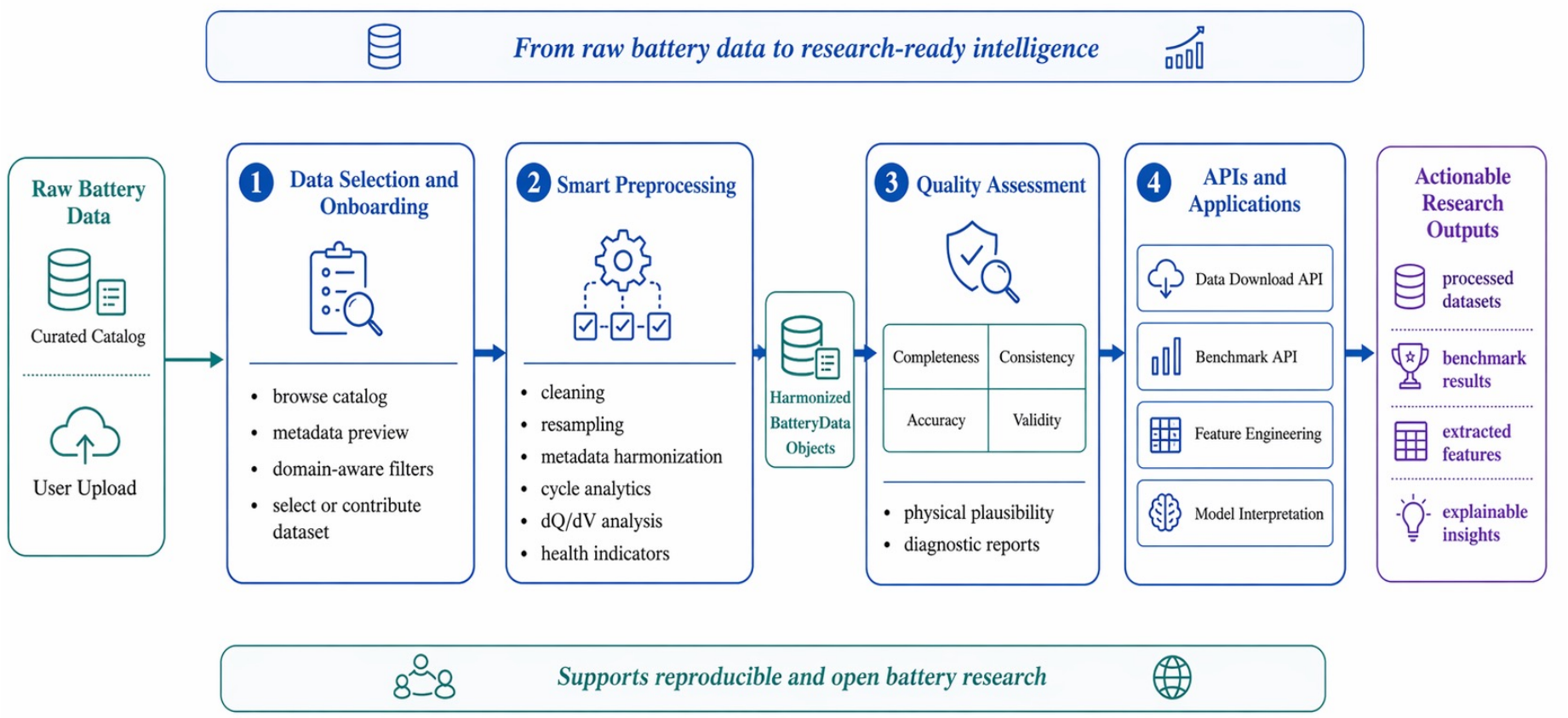}
\caption{The \system{} curation pipeline, from heterogeneous raw
sources to research-ready assets. Datasets are onboarded and
harmonized by LLM agents (Stages~1--2), screened by a validation gate
covering completeness, consistency, accuracy, and validity
(Stage~3)with physical-plausibility checks that generic validators
lack, and served through data, benchmark, feature, and
interpretation APIs (Stage~4); the resulting governed lake underpins
reproducible SOH/RUL benchmarking and open battery research.}
\label{fig:arch}
\end{figure*}

Fig.~\ref{fig:arch} shows the architecture of our BatteryLake. Raw sources enter through
two coupled agentic workflows, including metadata extraction
(\S\ref{sec:twostage}) and data conversion (\S\ref{sec:agentic}), 
whose outputs pass the selective human gate
(\S\ref{sec:hitl}) and the validation gate (\S\ref{sec:gate}) before
admission into the governed lake that feeds the benchmark
(Section~\ref{sec:lakehouse}).
 
\subsection{Two-Stage Provenance-Ranked Metadata Extraction}
\label{sec:twostage}
 
Metadata for a published dataset is dispersed across sources of
unequal authority: the repository landing page describes the released
artifact itself, while the companion paper describes the experiment
and may cover conditions or cells not included in the release. We
therefore extract in provenance order (Algorithm~\ref{alg:extract}).
Stage~A parses the dataset source page (OSF, Zenodo, Mendeley,
GitHub, or institutional pages) and attempts all fields in $F$.
Stage~B consults the paper only to fill fields that Stage~A left as
$\bot$ or supported with low confidence; a paper-derived value may
overwrite a source-derived one only if its confidence is strictly
higher, and each retained tuple keeps its provenance label $s_f$ so
that downstream users can distinguish artifact-level from
experiment-level facts. Conflicts between sources are not silently
resolved: both candidates are surfaced to the reviewer with their
evidence spans.
 
Two design rules are introduced to ensure reliable extraction. First, prompts require a verbatim quotation for every non-$\bot$ value, and a post-hoc string check rejects any tuple whose quotation does not occur in the source text, converting most hallucinations into detectable failures.
Second, the agent is explicitly forbidden from world-knowledge
imputation: it must not infer chemistry from a dataset name, form
factor from a familiar cell model, or protocol parameters from
community convention. 
This abstention strategy first prioritizes evidence fidelity over coverage. Stage~B then recovers part of the missing coverage. When permitted by the data license, supplementary files are also used to fill missing fields. No unsupported values are inferred.
 
\begin{algorithm}[t]
\caption{Two-stage evidence-grounded metadata extraction}
\label{alg:extract}
\begin{algorithmic}[1]
\Require fields $F$; source page $E_1$; paper $E_2$; threshold $\tau$
\Ensure tuples $\{(f, v_f, e_f, s_f, c_f)\}$, review queue $Q$
\State $\Phi_1 \gets \textsc{ExtractGrounded}(F, E_1)$
  \Comment{quote-or-abstain prompting}
\ForAll{$f \in F$ with $v_f \neq \bot$ in $\Phi_1$}
  \If{$e_f \not\sqsubseteq E_1$} set $v_f \gets \bot$
    \Comment{reject unverifiable evidence}
  \EndIf
\EndFor
\State $F' \gets \{f : v_f = \bot \ \text{or}\ c_f < \tau\}$
\State $\Phi_2 \gets \textsc{ExtractGrounded}(F', E_2)$;
  verify quotes against $E_2$
\ForAll{$f \in F'$}
  \If{$c_f^{(2)} > c_f^{(1)}$} adopt $\Phi_2[f]$ with $s_f = 2$
  \ElsIf{both $\neq \bot$ and values disagree} flag conflict for review
  \EndIf
\EndFor
\State $Q \gets \{f : c_f < \tau \ \text{or flagged}\}$;
  auto-accept the rest
\State \Return tuples, $Q$
\end{algorithmic}
\end{algorithm}
 
\subsection{Agentic Conversion: Classify, Synthesize, Validate,
Repair}\label{sec:agentic}
 
Raw measurement files cannot be handled by a fixed parser bank. 
Across our catalog we observe 18 distinct layout families in
CSV alone, besides MAT structures with nested cycle records, Excel
workbooks with per-cell sheets, and vendor exports. \system{}
instead packages the canonical schema, the dataset name provided by
the contributor, and the executable validator into a self-contained
\emph{curation skill} that an LLM agent executes locally against the
raw folder (Algorithm~\ref{alg:convert}); raw data never leaves the
contributor's machine, which also sidesteps licensing and volume
constraints of server-side ingestion.
 
The agent first \emph{inspects} the source, including file listings, headers,
MAT keys, sample rows, and classifies each file's role $\rho$;
DataCite JSONs, READMEs, and PDFs are diverted to the extraction
workflow as evidence. It then proposes the grounded attribute mapping
$h$ with unit conversions, synthesizes a dataset-specific converter
$\pi$, and executes it. The validator replays $\Sigma$ on the outputs
and emits a machine-readable report; on failure, the error report is
fed back and the agent repairs $\pi$, up to $k$ rounds. This
generate--validate--repair loop replaces trust in the agent with
trust in the acceptance test: an incorrect converter can waste
attempts but cannot silently admit malformed data.
 
\begin{algorithm}[t]
\caption{Converter synthesis with validation-guided repair}
\label{alg:convert}
\begin{algorithmic}[1]
\Require raw files $R$; schema and constraints $\Sigma$; budget $k$
\Ensure $(\hat{T}, \hat{S}, \hat{M_c})$ or failure report
\State $\rho \gets \textsc{ClassifyRoles}(R)$;
  route \textsf{metadata\_only} files to Algorithm~\ref{alg:extract}
\State $h \gets \textsc{GroundedMapping}(R_\textsf{meas})$
  \Comment{bind columns by header, unit, statistics; else abstain}
\State human confirms/edits $\rho, h$ \Comment{selective gate,
  \S\ref{sec:hitl}}
\For{$t = 1 \ldots k$}
  \State $\pi_t \gets \textsc{SynthesizeConverter}(R_\textsf{meas},
    h, \Sigma)$
  \State $(\hat{T}, \hat{S}, \hat{M_c}) \gets \pi_t(R_\textsf{meas})$;
   \ $\mathit{rep} \gets \textsc{Validate}(\hat{T}, \hat{S},
    \hat{M_c}, \Sigma)$
  \If{$\mathit{rep}.\textsf{pass}$} \Return outputs with
    $\mathit{rep}$ \EndIf
  \State feed $\mathit{rep}$ back to the agent for repair
\EndFor
\State \Return failure with last $\mathit{rep}$
\end{algorithmic}
\end{algorithm}

\begin{figure*}[t]
\centering
\includegraphics[width=0.66\textwidth]{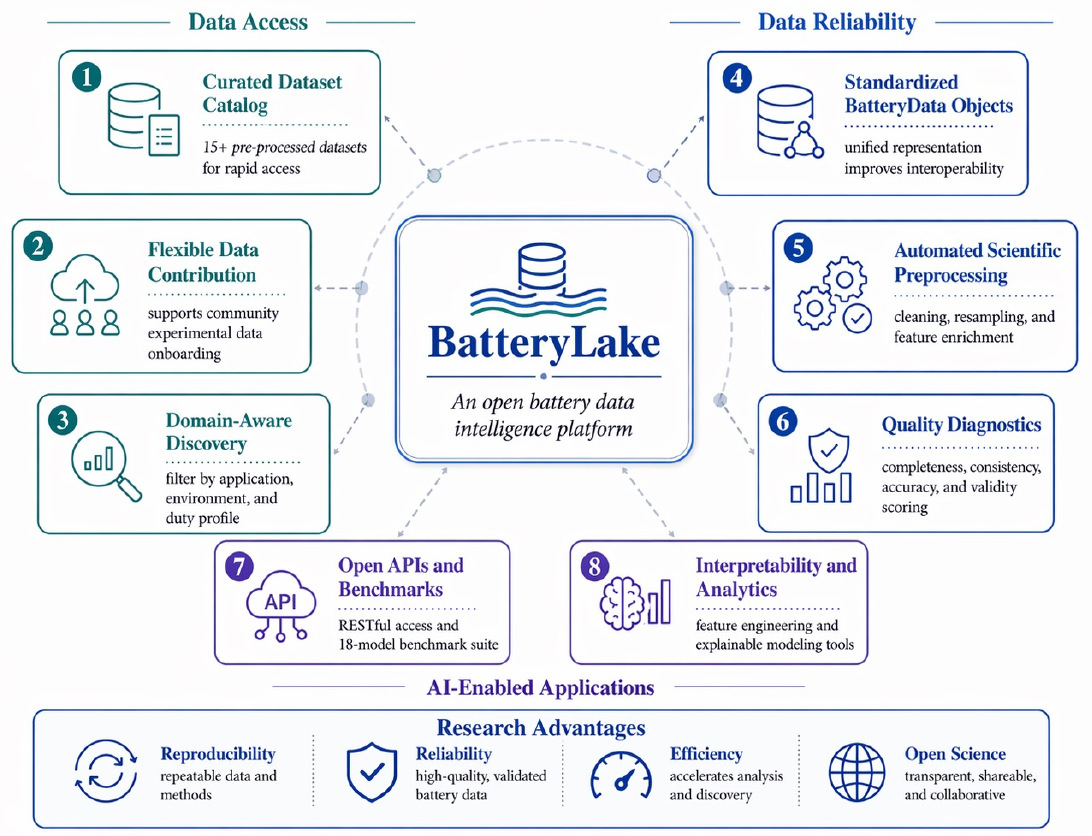}
\caption{Capability overview of the \system{} platform. Eight
functional pillars span curated data access, data reliability, and
AI-enabled applications, and together deliver reproducible,
reliable, and openly shareable battery data for downstream research.}
\label{fig:platform}
\end{figure*}

\subsection{The Validation Gate}\label{sec:gate}

The gate evaluates $|\Sigma| = 26$ rules grouped into four scored
dimensions, producing a per-dataset diagnostic report that can block
or admit downstream use:
\begin{itemize}
\item \emph{Validity}: schema compliance, which requireds columns, types,
unit-encoded names, key integrity across the four layers;
\item \emph{Completeness}: coverage of required channels and cycles,
missingness rates per signal;
\item \emph{Consistency}: monotonic timestamps, contiguous cycle
numbering, a single current sign convention, step-type coherence;
\item \emph{Accuracy (physical plausibility)}: voltage within
chemistry-consistent bounds, per-cycle coulombic efficiency within
$[95\%, 105\%]$, capacity-fade trajectories non-increasing up to
bounded recovery windows, cell temperature within a tolerance of the
stated condition, charge/discharge energy balance.
\end{itemize}
Generic validation systems~\cite{schelter2018} cover the first three dimensions. The fourth encodes electrochemical domain knowledge, and in practice it is the dimension that catches unit errors and sign-convention bugs: values that pass schema validation but are physically impossible. Each dataset's gate score and rule-level outcomes are stored alongside the data, so benchmark users can filter on curation quality rather than taking it on faith.
 
\subsection{Selective Human Verification in Practice}
\label{sec:hitl}
 
The reviewer interface presents each pending decision as a row
(field, value, verbatim evidence, provenance, confidence); the
reviewer edits or confirms rows and must tick a final attestation
before export, so provenance of \emph{human} decisions is recorded
just like machine ones. Confidence thresholds are set per field
group: identity-critical fields (chemistry, nominal capacity) use a
conservative $\tau$ so they are almost always reviewed, while
low-stakes descriptive fields tolerate higher auto-acceptance. The
same gate covers conversion: role labels $\rho$ and mappings $h$ are
confirmed before any converter runs, because a wrong mapping is far
cheaper to fix before synthesis than after. 
%Section~\ref{sec:eval} quantifies the resulting effort--risk trade-off.
 
\section{Governed Lakehouse and Open Benchmark}\label{sec:lakehouse}
 
The curated pipeline of Section~\ref{sec:framework} materializes as a
governed lakehouse whose capabilities are summarized in
Fig.~\ref{fig:platform}: curated access (catalog, community
contribution, domain-aware discovery), reliability (standardized data
objects, automated preprocessing, quality diagnostics), and
AI-enabled applications (open APIs and benchmarks, interpretability).
These capabilities operationalize the four design goals of
reproducibility, reliability, efficiency, and open science.

\subsection{Curated Collection}
 
\begin{table}[t]
\centering
\caption{\system{} catalog snapshot at submission. ``Curated''
datasets have passed all four ETL stages (metadata, time series,
cycle summary, QC); the registry additionally tracks datasets queued
for onboarding.}
\label{tab:catalog}
\begin{tabular}{@{}>{\raggedright\arraybackslash}p{0.63\columnwidth}
                   >{\raggedleft\arraybackslash}p{0.30\columnwidth}@{}}
\toprule
Registered datasets & 41 \\
Fully curated datasets & 12 \\
Source institutions (curated) & 12 \\
Cells (curated subset) & $\sim$720 \\
Charge--discharge cycles (curated) & $\sim$323K \\
Data volume (curated) & $\sim$9.5\,GB \\
Chemistries & LFP, NMC, LCO, NCA, multi \\
Form factors & 18650, 21700, pouch, prismatic \\
Publication span & 2007--2026 \\
\bottomrule
\end{tabular}
\end{table}
 
\begin{figure}[t]
\centering
\includegraphics[width=\columnwidth]{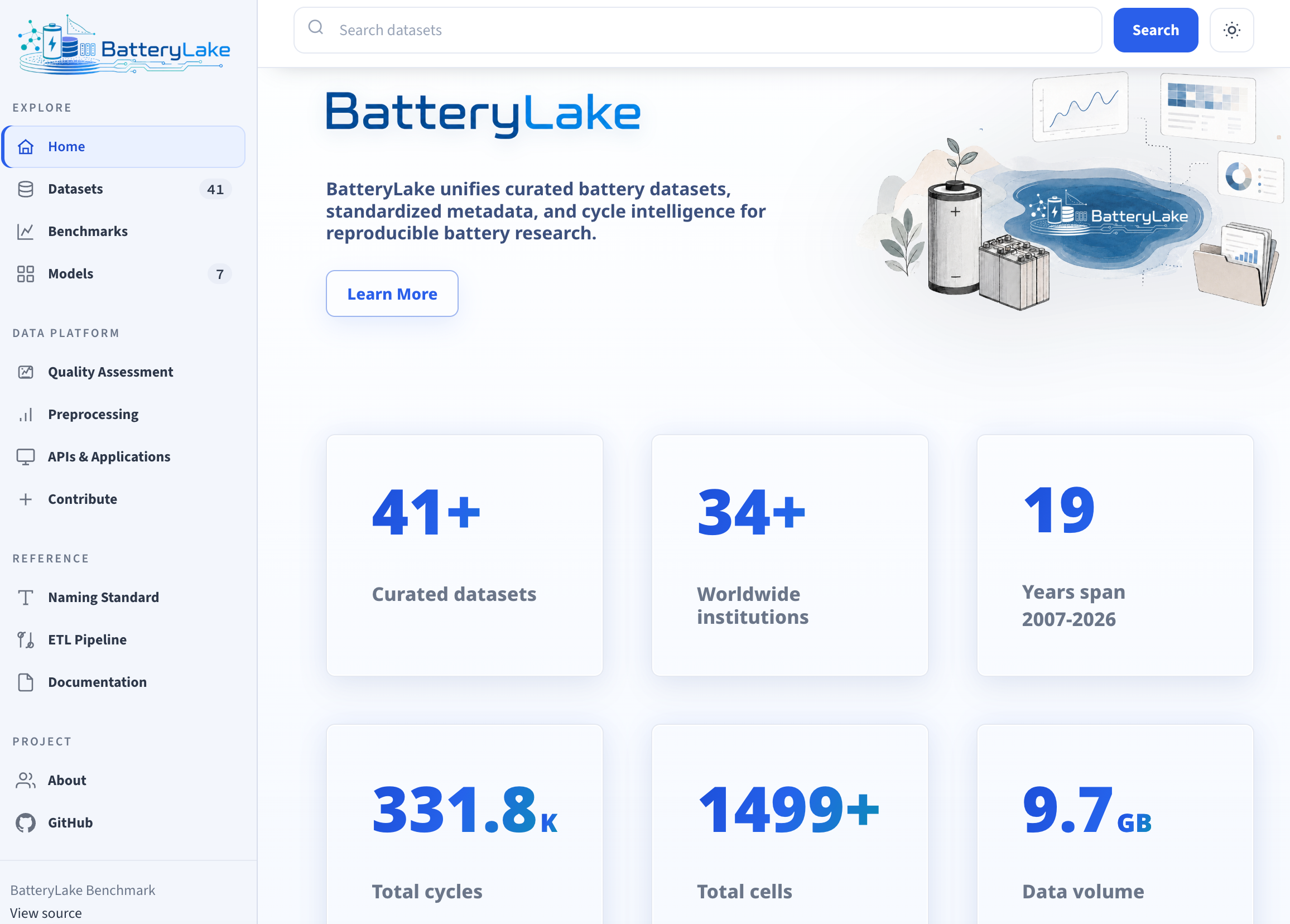}
\caption{The \system{} landing page. A left-hand navigation exposes
dataset browsing, benchmarks, the model library, and the data-platform
services (quality assessment, preprocessing, APIs), while headline
cards report the aggregate scope of the curated collection, including number of datasets, contributing institutions, cells, charge--discharge cycles, data volume, and publication span.}
\label{fig:home}
\end{figure}

Table~\ref{tab:catalog} summarizes the collection, which spans cycle
aging, safety/abuse, impedance, and EV field categories, and includes
the canonical corpora (NASA PCoE~\cite{saha2007},
CALCE~\cite{xing2013}, Severson et al.~\cite{severson2019},
Oxford~\cite{birkl2017}) alongside long-tail releases spanning
12 laboratories worldwide. Every curated dataset ships the four-layer
object of \S\ref{sec:model} plus a \emph{processing manifest}
recording raw-file inventory, mapping rules, unit conversions, cycle
segmentation logic, label definitions, and validator version, making
each dataset a versioned research artifact whose benchmark numbers
can be traced back to raw bytes. Fig.~\ref{fig:home} shows the
platform landing page, which exposes the aggregate scope of the
collection and provides entry points to browse, filter, and
contribute datasets.

\subsection{Benchmark Design}
 
\begin{figure}[t]
\centering
\includegraphics[width=\columnwidth]{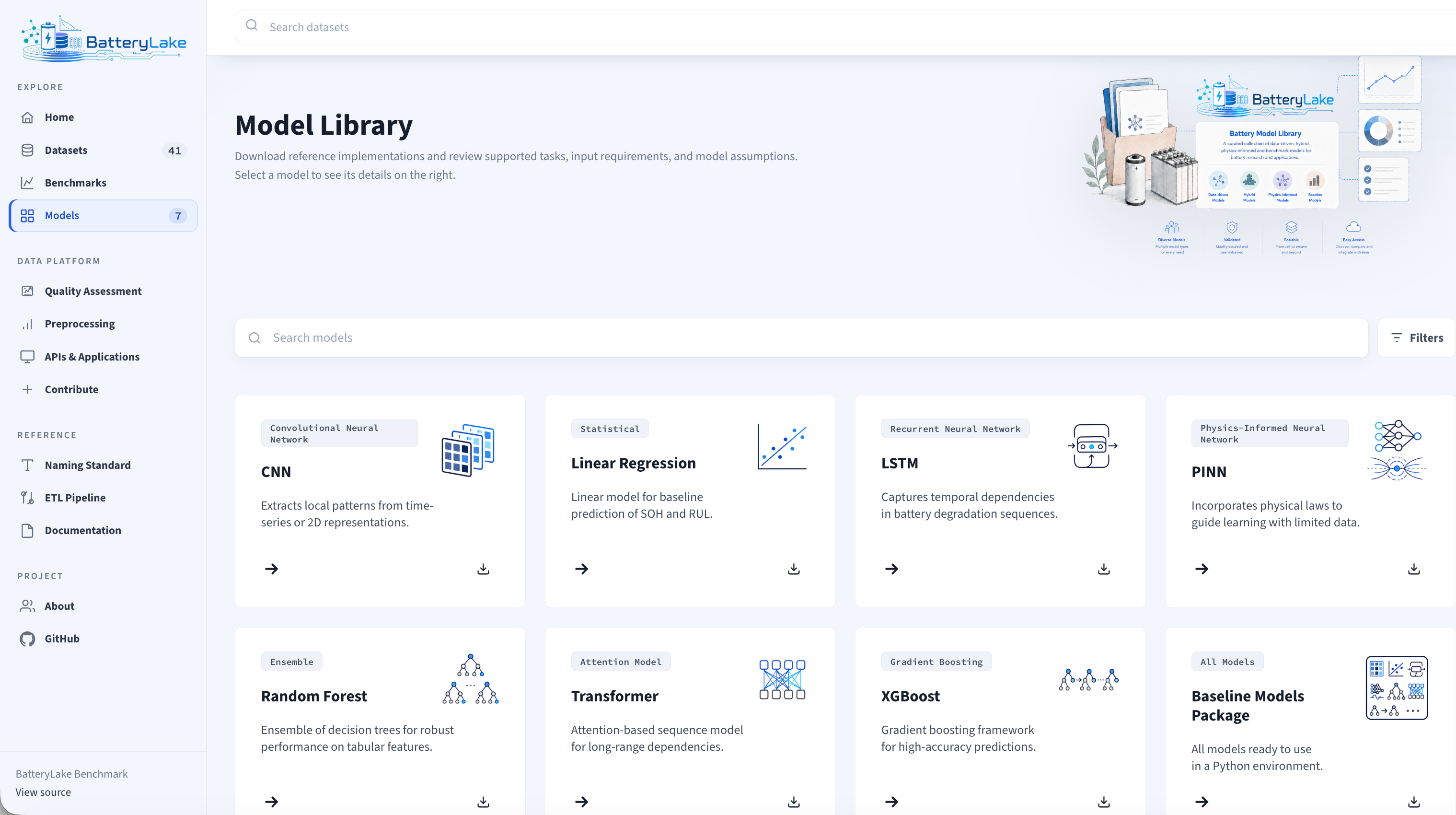}
\caption{The \system{} model library. Each baseline family---spanning
statistical (linear regression), ensemble (random forest, XGBoost),
deep-learning (MLP, CNN, LSTM, Transformer), and physics-informed
(PINN) models---ships a downloadable reference implementation with
documented supported tasks, input requirements, and modeling
assumptions, so that benchmark comparisons use shared model code.}
\label{fig:models}
\end{figure}

\begin{figure*}[t]
\centering
\begin{subfigure}[t]{0.32\textwidth}
  \includegraphics[width=\textwidth]{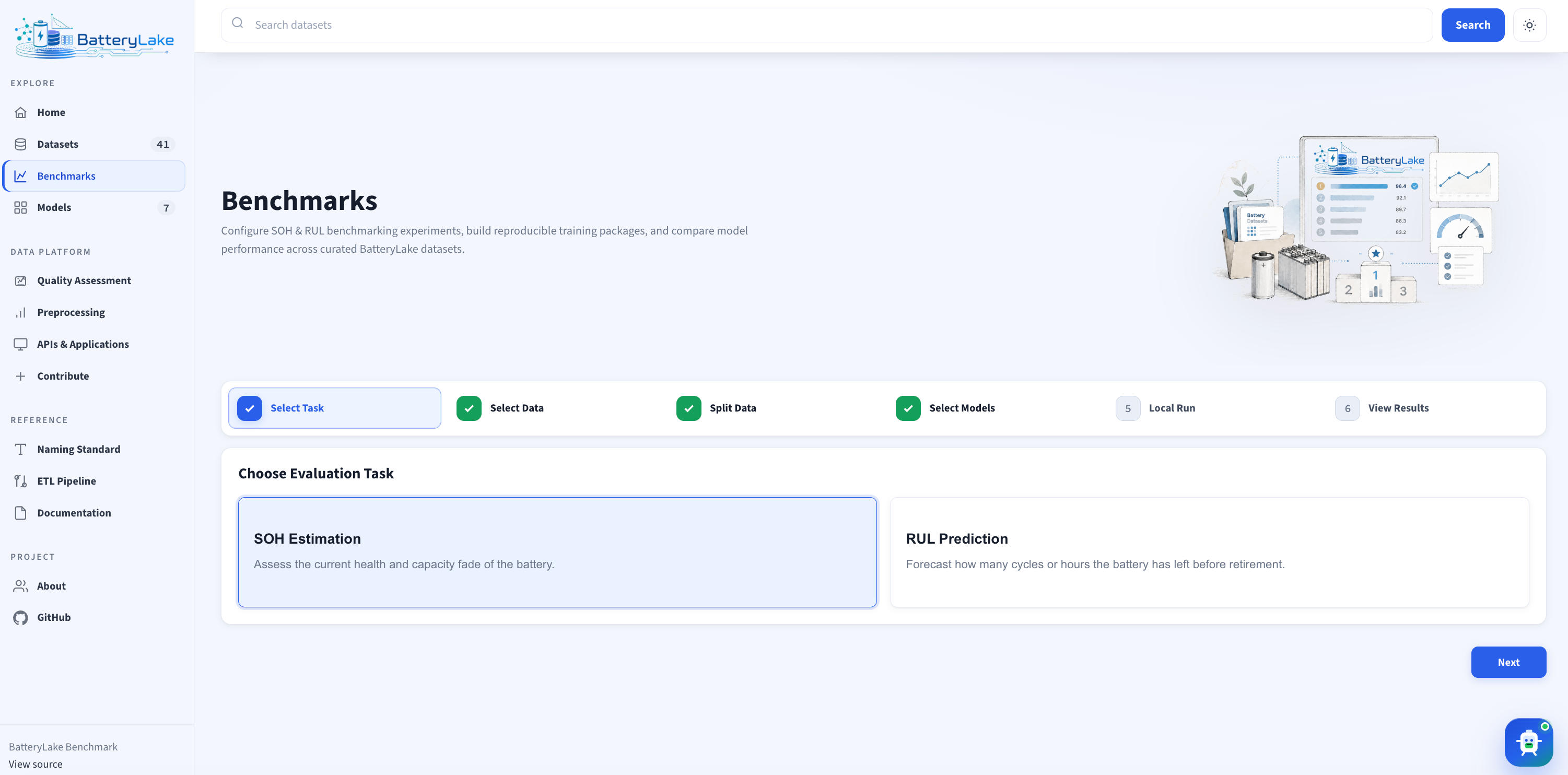}
  \caption{Select Task}
  \label{fig:bench-task}
\end{subfigure}
\hfill
\begin{subfigure}[t]{0.32\textwidth}
  \includegraphics[width=\textwidth]{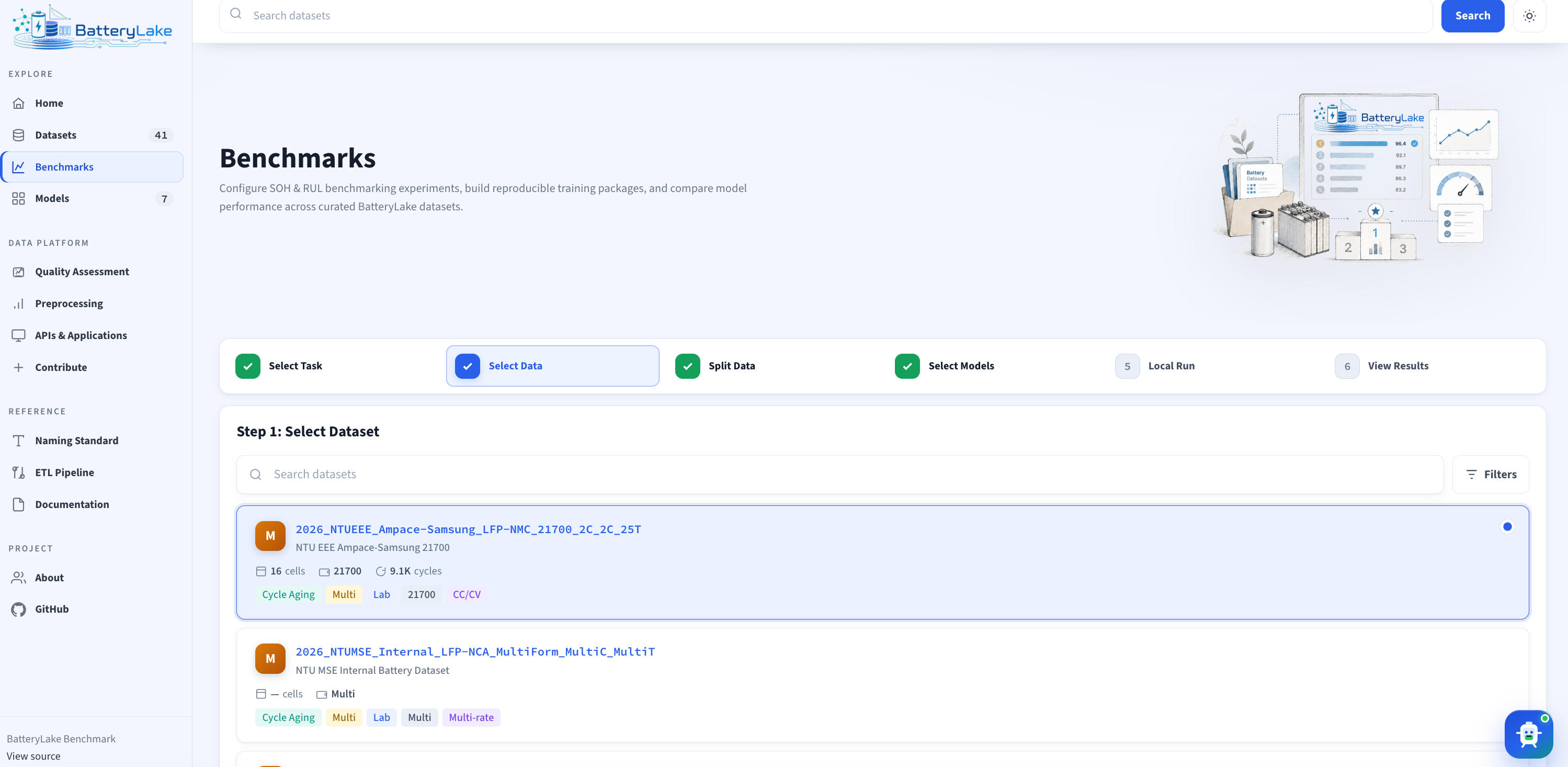}
  \caption{Select Data}
  \label{fig:bench-data}
\end{subfigure}
\hfill
\begin{subfigure}[t]{0.32\textwidth}
  \includegraphics[width=\textwidth]{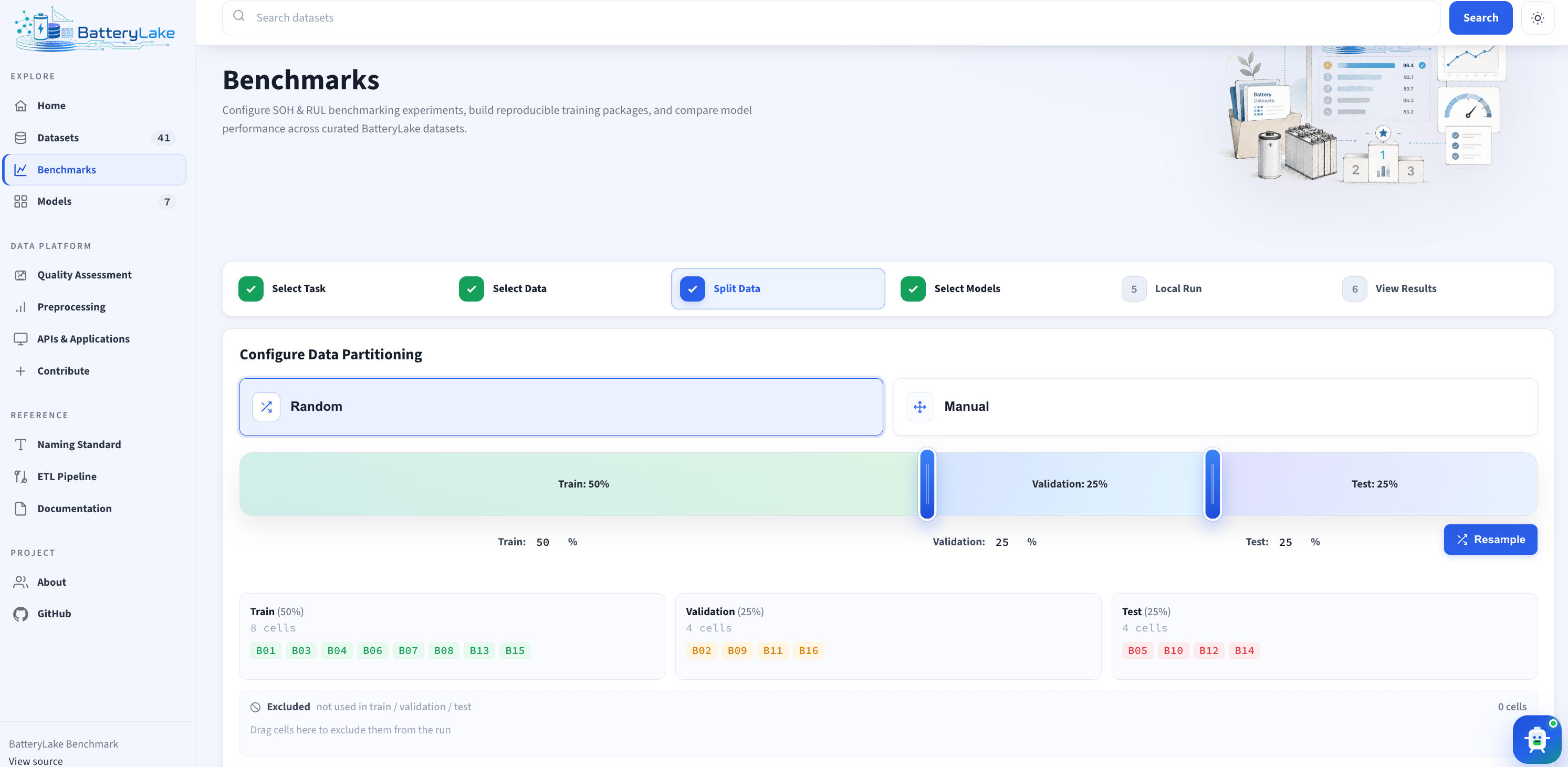}
  \caption{Split Data}
  \label{fig:bench-split}
\end{subfigure}

\vspace{2mm}

\begin{subfigure}[t]{0.32\textwidth}
  \includegraphics[width=\textwidth]{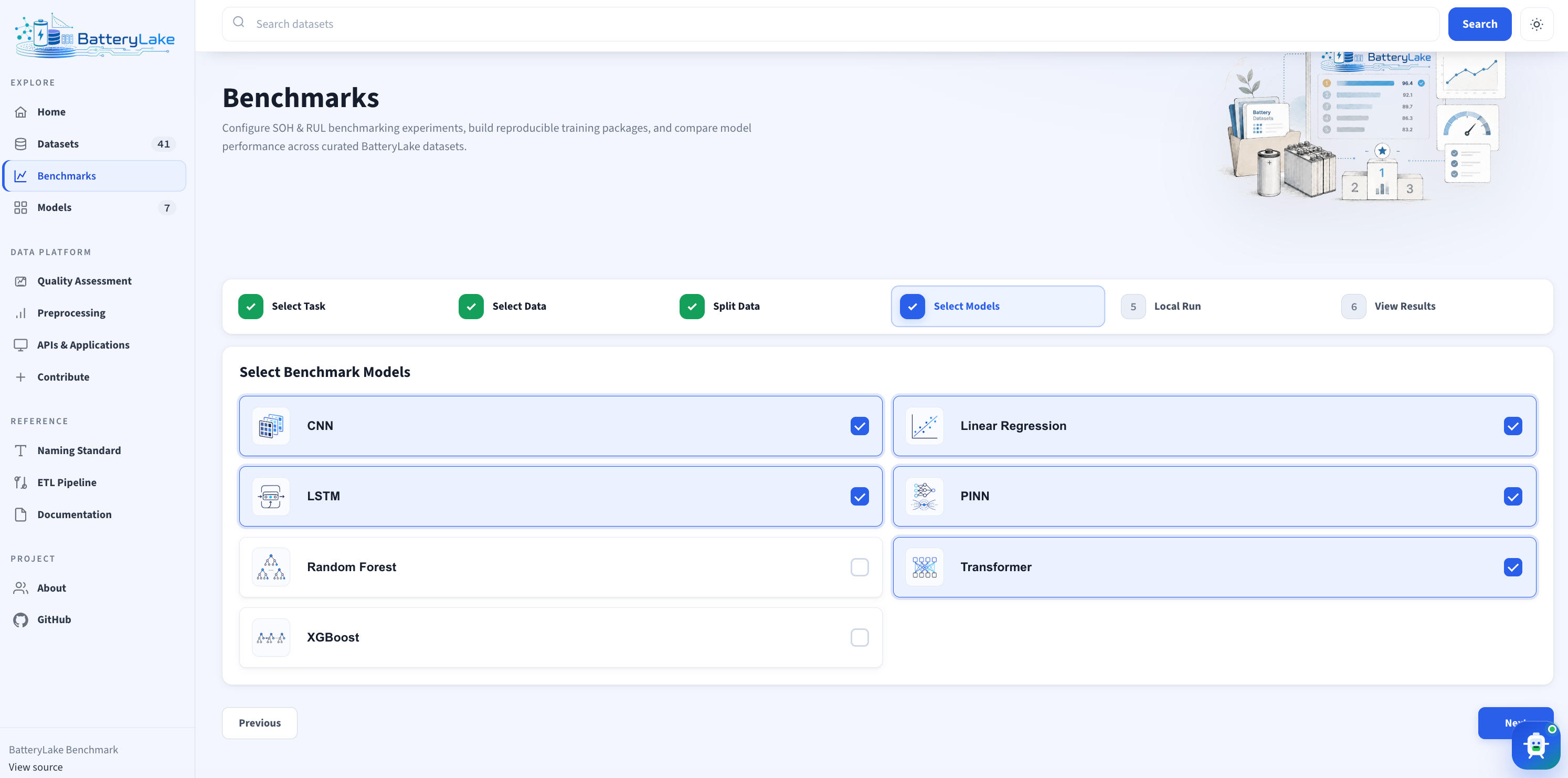}
  \caption{Select Models}
  \label{fig:bench-models}
\end{subfigure}
\hfill
\begin{subfigure}[t]{0.32\textwidth}
  \includegraphics[width=\textwidth]{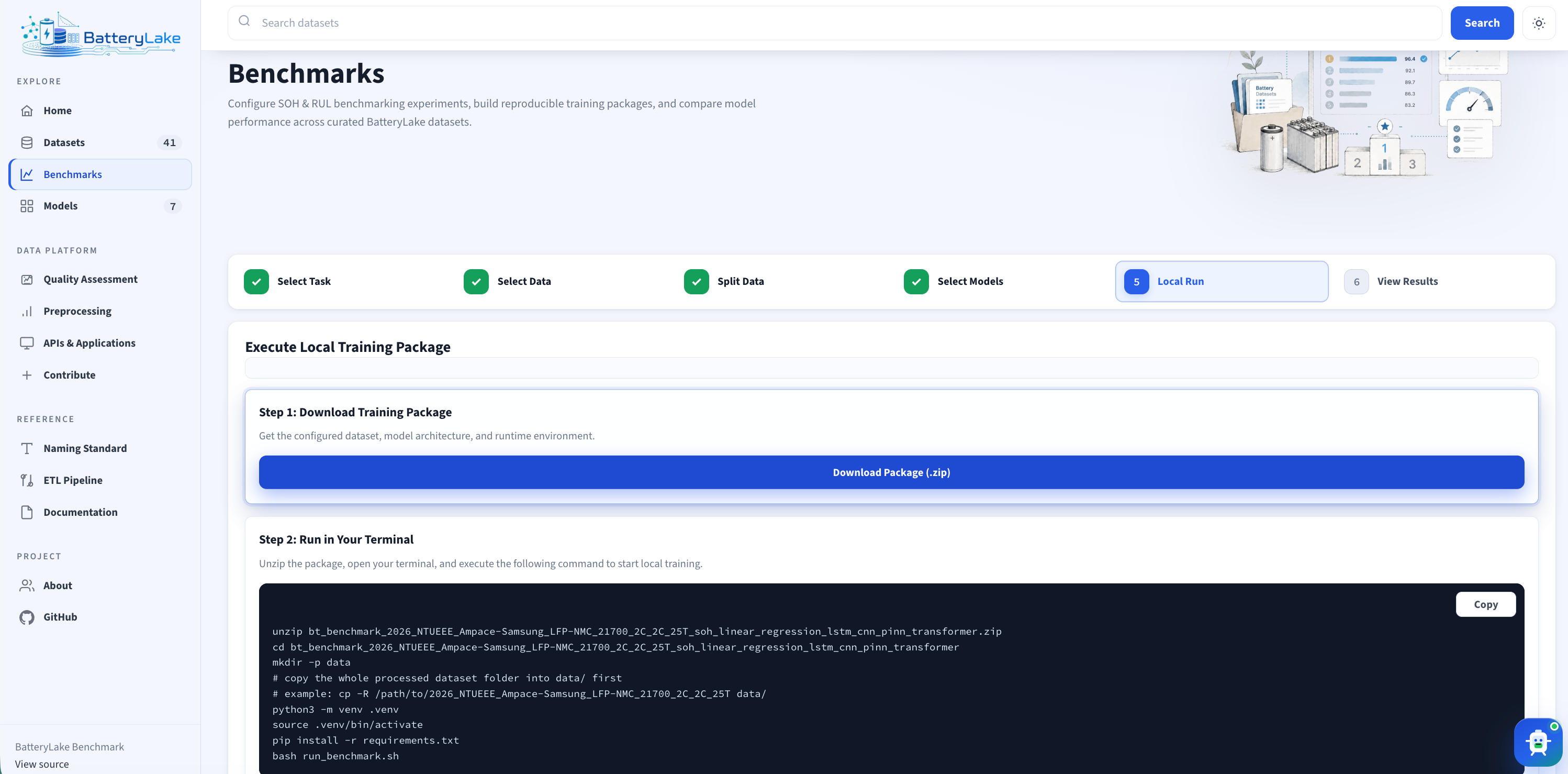}
  \caption{Local Run}
  \label{fig:bench-run}
\end{subfigure}
\hfill
\begin{subfigure}[t]{0.32\textwidth}
  \includegraphics[width=\textwidth]{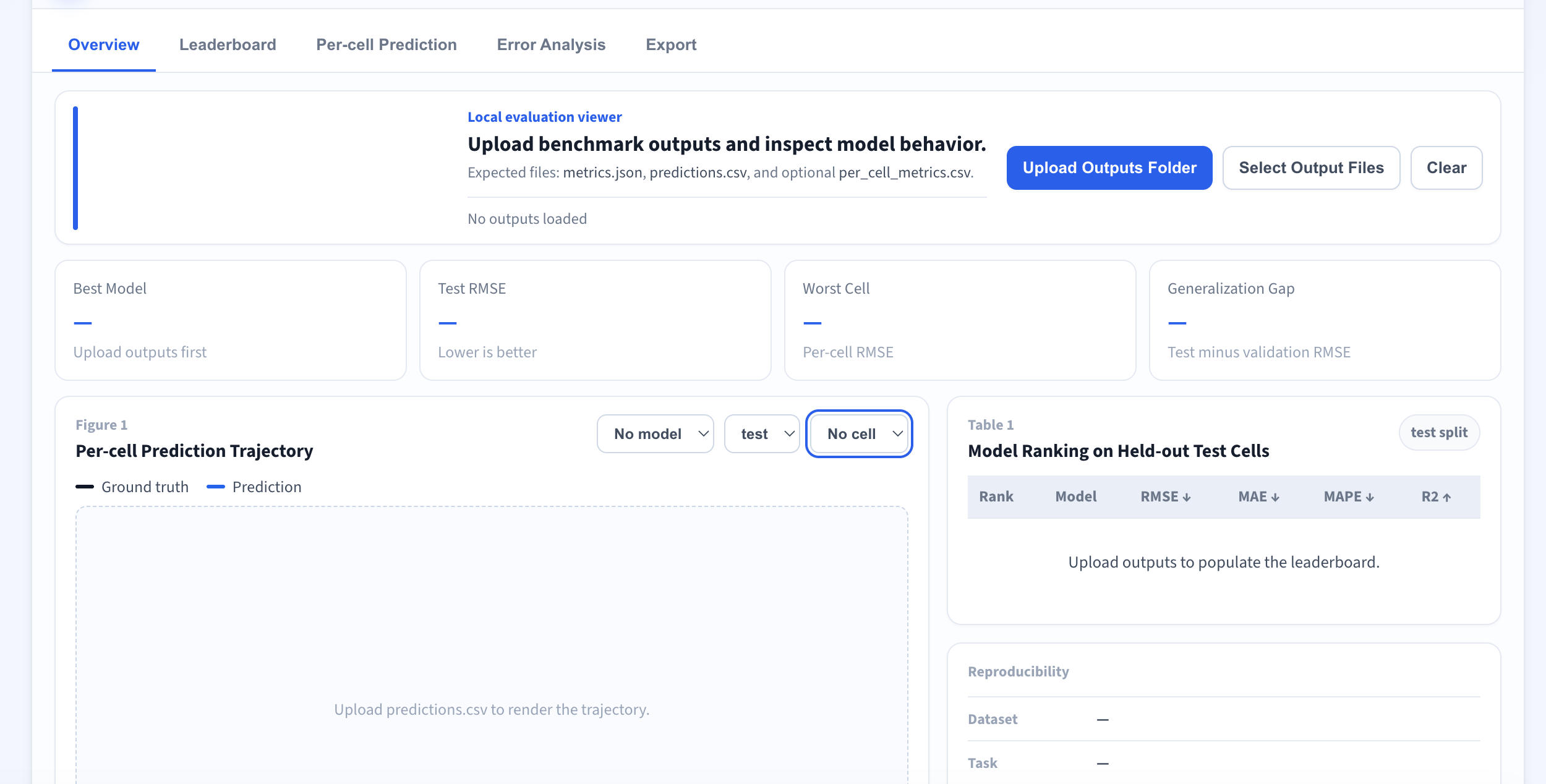}
  \caption{View Results}
  \label{fig:bench-results}
\end{subfigure}
\caption{The \system{} benchmark configuration workflow, shown as a six-step wizard. (a) \emph{Select Task}: choose SOH estimation or RUL prediction. (b) \emph{Select Data}: pick curated datasets, each shown with chemistry, cell count, and cycle count. (c) \emph{Split Data}: partition cells into train, validation, and test, at random or by hand. (d) \emph{Select Models}: choose baseline families to run. (e) \emph{Local Run}: export a self-contained training package and run it locally with the printed commands. (f) \emph{View Results}: inspect per-model metrics and compare performance across the selected models. Every choice is recorded, so the reported metrics are traceable and reproducible.}
\label{fig:benchmark}
\end{figure*}

On the curated lake, \system{} defines two primary tasks with
standardized labels: \emph{SOH estimation} (predict capacity
retention at a cycle from partial-cycle signals or cycle-level
features) and \emph{RUL prediction} (predict cycles remaining to a
manifest-recorded end-of-life threshold). Because split design
dominates apparent performance in battery ML, every experiment
declares one of three protocols: \emph{random} (cycles shuffled;
optimistic), \emph{temporal} (early cycles train, late cycles test;
forecasting under distribution shift), and \emph{cross-cell} (whole
cells held out; generalization to unseen units). Eight baseline
families are provided, which are linear regression, random forest, XGBoost, MLP, CNN, LSTM, Transformer, and physics-informed neural networks (PINN) with fixed configurations, seeds, and RMSE/MAE/MAPE/$R^2$ metrics. Rather than executing training server-side, the platform
exports a reproducible \emph{training package} (config, split
assignment, feature schema, dataloader, model code, run script) that
users run locally and whose outputs the platform's evaluation viewer
ingests for leaderboard comparison; the package pins every choice
that normally varies silently between papers. Fig.~\ref{fig:models}
shows the model library from which these baselines are drawn, and
Fig.~\ref{fig:benchmark} shows the guided workflow that assembles a
benchmark run from dataset and cell selection, feature configuration,
and split assignment through to exported metrics.

\section{Discussion and Limitations}\label{sec:discussion}
 
\textbf{Scope of guarantees.} The residual-risk bound of
\S\ref{sec:selective-def} conditions on correct reviewer decisions
and calibrated confidences; we report calibration error and
inter-annotator agreement so users can judge both assumptions. The
grounding constraint prevents unsupported values but cannot detect a
source page that is itself wrong; provenance labels at least make
such errors attributable. \textbf{Evidence access.} Some companion
papers are paywalled and some metadata exists only in supplementary
archives; our abstention-first design degrades gracefully (coverage
drops, fidelity does not), but coverage numbers should be read with
this in mind. \textbf{Local agent execution.} Running converters on
the contributor's machine preserves privacy and licenses but means
the platform verifies \emph{reports} of validation rather than
re-executing it; hash-pinned validators and replayable manifests
mitigate, and server-side re-validation for redistributable datasets
is future work. \textbf{Generality.} The four-layer model and the
26-rule gate are battery-specific, but the framework's contract of 
quote-or-abstain extraction, selective verification, validator-guided
synthesis, which transfers to other instrument-generated scientific data; adapting the constraint set is the only domain-specific step.
 
\section{Conclusion}\label{sec:conclusion}
 
We presented \system{}, a governed lakehouse and open benchmark that
converts fragmented public battery aging data into reproducible
research assets through agentic, evidence-grounded curation. By
formalizing onboarding as grounded extraction with abstention plus
validator-checked converter synthesis, and by casting human review as
selective prediction, \system{} replaces trust in either humans or
models with an auditable pipeline whose residual risk is measured and
tunable. The released benchmark with $41$ datasets, standardized
SOH/RUL tasks, split protocols, and manifest-pinned baselines offers the battery ML community a common, provenance-complete substrate, and the curation framework offers the data engineering community a template for governing the long tail of scientific data.
 
\bibliographystyle{IEEEtran}

\begin{thebibliography}{10}
 
\bibitem{severson2019}
K.~A. Severson \emph{et~al.}, ``Data-driven prediction of battery
cycle life before capacity degradation,'' \emph{Nature Energy},
vol.~4, no.~5, pp. 383--391, 2019.
 
\bibitem{attia2020}
P.~M. Attia \emph{et~al.}, ``Closed-loop optimization of
fast-charging protocols for batteries with machine learning,''
\emph{Nature}, vol. 578, no. 7795, pp. 397--402, 2020.
 
\bibitem{saha2007}
B.~Saha and K.~Goebel, ``Battery data set,'' NASA Ames Prognostics
Data Repository, NASA Ames Research Center, Moffett Field, CA, 2007.
 
\bibitem{xing2013}
Y.~Xing, E.~W.~M. Ma, K.-L. Tsui, and M.~Pecht, ``An ensemble model
for predicting the remaining useful performance of lithium-ion
batteries,'' \emph{Microelectronics Reliability}, vol.~53, no.~6,
pp. 811--820, 2013.
 
\bibitem{birkl2017}
C.~R. Birkl, ``Oxford battery degradation dataset 1,'' University of
Oxford, 2017.
 
\bibitem{dosreis2021}
G.~dos Reis, C.~Strange, M.~Yadav, and S.~Li, ``Lithium-ion battery
data and where to find it,'' \emph{Energy and AI}, vol.~5, p. 100081,
2021.
 
\bibitem{zhang2024batteryml}
X.~Zhang \emph{et~al.}, ``{BatteryML}: An open-source platform for
machine learning on battery degradation,'' in \emph{Proc. Int. Conf.
Learning Representations (ICLR)}, 2024.
 
\bibitem{herring2020}
P.~Herring \emph{et~al.}, ``{BEEP}: A python library for battery
evaluation and early prediction,'' \emph{SoftwareX}, vol.~11,
p. 100506, 2020.
 
\bibitem{clark2022}
S.~Clark \emph{et~al.}, ``Toward a unified description of battery
data,'' \emph{Advanced Energy Materials}, vol.~12, no.~17,
p. 2102702, 2022.
 
\bibitem{madhavan2001}
J.~Madhavan, P.~A. Bernstein, and E.~Rahm, ``Generic schema matching
with {Cupid},'' in \emph{Proc. 27th Int. Conf. Very Large Data Bases
(VLDB)}, 2001, pp. 49--58.
 
\bibitem{li2020ditto}
Y.~Li, J.~Li, Y.~Suhara, A.~Doan, and W.-C. Tan, ``Deep entity
matching with pre-trained language models,'' \emph{Proc. VLDB
Endow.}, vol.~14, no.~1, pp. 50--60, 2020.
 
\bibitem{tu2023unicorn}
J.~Tu \emph{et~al.}, ``Unicorn: A unified multi-tasking model for
supporting matching tasks in data integration,'' \emph{Proc. ACM
Manag. Data (SIGMOD)}, vol.~1, no.~1, pp. 84:1--84:26, 2023.
 
\bibitem{narayan2022}
A.~Narayan, I.~Chami, L.~Orr, and C.~R{\'e}, ``Can foundation models
wrangle your data?'' \emph{Proc. VLDB Endow.}, vol.~16, no.~4,
pp. 738--746, 2022.
 
\bibitem{arora2023}
S.~Arora \emph{et~al.}, ``Language models enable simple systems for
generating structured views of heterogeneous data lakes,''
\emph{Proc. VLDB Endow.}, vol.~17, no.~2, pp. 92--105, 2023.
 
\bibitem{zhang2024jellyfish}
H.~Zhang, Y.~Dong, C.~Xiao, and M.~Oyamada, ``Jellyfish: Instruction-
tuning local large language models for data preprocessing,'' in
\emph{Proc. Conf. Empirical Methods in Natural Language Processing
(EMNLP)}, 2024, pp. 8754--8782.
 
\bibitem{yao2023react}
S.~Yao \emph{et~al.}, ``{ReAct}: Synergizing reasoning and acting in
language models,'' in \emph{Proc. Int. Conf. Learning Representations
(ICLR)}, 2023.
 
\bibitem{nargesian2019}
F.~Nargesian, E.~Zhu, R.~J. Miller, K.~Q. Pu, and P.~C. Arocena,
``Data lake management: Challenges and opportunities,'' \emph{Proc.
VLDB Endow.}, vol.~12, no.~12, pp. 1986--1989, 2019.
 
\bibitem{armbrust2020}
M.~Armbrust \emph{et~al.}, ``Delta lake: High-performance {ACID}
table storage over cloud object stores,'' \emph{Proc. VLDB Endow.},
vol.~13, no.~12, pp. 3411--3424, 2020.
 
\bibitem{schelter2018}
S.~Schelter, D.~Lange, P.~Schmidt, M.~Celikel, F.~Biessmann, and
A.~Grafberger, ``Automating large-scale data quality verification,''
\emph{Proc. VLDB Endow.}, vol.~11, no.~12, pp. 1781--1794, 2018.
 
\bibitem{rekatsinas2017}
T.~Rekatsinas, X.~Chu, I.~F. Ilyas, and C.~R{\'e}, ``{HoloClean}:
Holistic data repairs with probabilistic inference,'' \emph{Proc.
VLDB Endow.}, vol.~10, no.~11, pp. 1190--1201, 2017.
 
\bibitem{elyaniv2010}
R.~El-Yaniv and Y.~Wiener, ``On the foundations of noise-free
selective classification,'' \emph{J. Mach. Learn. Res.}, vol.~11,
pp. 1605--1641, 2010.
 
\bibitem{geifman2017}
Y.~Geifman and R.~El-Yaniv, ``Selective classification for deep
neural networks,'' in \emph{Advances in Neural Information Processing
Systems (NeurIPS)}, 2017, pp. 4878--4887.
 
\bibitem{guo2017}
C.~Guo, G.~Pleiss, Y.~Sun, and K.~Q. Weinberger, ``On calibration of
modern neural networks,'' in \emph{Proc. 34th Int. Conf. Machine
Learning (ICML)}, 2017, pp. 1321--1330.
 
\bibitem{wilkinson2016}
M.~D. Wilkinson \emph{et~al.}, ``The {FAIR} guiding principles for
scientific data management and stewardship,'' \emph{Scientific Data},
vol.~3, p. 160018, 2016.
 
\end{thebibliography}

\end{document}